\documentclass[letterpaper]{article} 
\usepackage{aaai25}  
\usepackage{times}  
\usepackage{helvet}  
\usepackage{courier}  
\usepackage[hyphens]{url}  
\usepackage{graphicx} 
\usepackage{natbib}  
\usepackage{caption} 
\usepackage{algorithm}
\usepackage{algorithmic}

\usepackage{newfloat}
\usepackage{listings}
\DeclareCaptionStyle{ruled}{labelfont=normalfont,labelsep=colon,strut=off} 
\floatstyle{ruled}
\newfloat{listing}{tb}{lst}{}
\floatname{listing}{Listing}
\usepackage{tikz}
\usepackage{amssymb}
\usepackage{utfsym}
\usepackage{amsmath}
\usepackage{subcaption}

\title{Towards Efficient and Intelligent Laser Weeding: Method and Dataset for Weed Stem Detection}

\author {
    Dingning Liu\textsuperscript{\rm 1}\equalcontrib,
    Jinzhe Li\textsuperscript{\rm 1}\equalcontrib,
    Haoyang Su\textsuperscript{\rm 1}\equalcontrib,
    Bei Cui\textsuperscript{\rm 2}\thanks{Corresponding author.},
    Zhihui Wang\textsuperscript{\rm 3},
    Qingbo Yuan\textsuperscript{\rm 2},\\
    Wanli Ouyang\textsuperscript{\rm 1},
    Nanqing Dong\textsuperscript{\rm 1}\footnotemark[2]
}

\affiliations {
    \textsuperscript{\rm 1}Shanghai Artificial Intelligence Laboratory\\
    \textsuperscript{\rm 2}Yazhouwan National Laboratory\\
    \textsuperscript{\rm 3}Dalian University of Technology\\
\{liudingning, dongnanqing\}@pjlab.org.cn
}

\begin{document}
\maketitle

\begin{abstract}
Weed control is a critical challenge in modern agriculture, as weeds compete with crops for essential nutrient resources, significantly reducing crop yield and quality. Traditional weed control methods, including chemical and mechanical approaches, have real-life limitations such as associated environmental impact and efficiency. 
An emerging yet effective approach is laser weeding, which uses a laser beam as the stem cutter. Although there have been studies that use deep learning in weed recognition, its application in intelligent laser weeding still requires a comprehensive understanding. Thus, this study represents the first empirical investigation of weed recognition for laser weeding. To increase the efficiency of laser beam cut and avoid damaging the crops of interest, the laser beam shall be directly aimed at the weed root. Yet, weed stem detection remains an under-explored problem. We integrate the detection of crop and weed with the localization of weed stem into one end-to-end system. To train and validate the proposed system in a real-life scenario, we curate and construct a high-quality weed stem detection dataset with human annotations. The dataset consists of 7,161 high-resolution pictures collected in the field with annotations of 11,151 instances of weed. 
Experimental results show that the proposed system improves weeding accuracy by 6.7\% and reduces energy cost by 32.3\% compared to existing weed recognition systems.

\end{abstract}

\begin{links}
\link{Code \& Dataset}{https://github.com/open-sciencelab/WeedStemDetection}
\end{links}

\begin{figure}[!ht]
    \centering
   
    \begin{subfigure}[b]{0.48\linewidth}
        \centering
        \includegraphics[height=4cm,width=\textwidth]{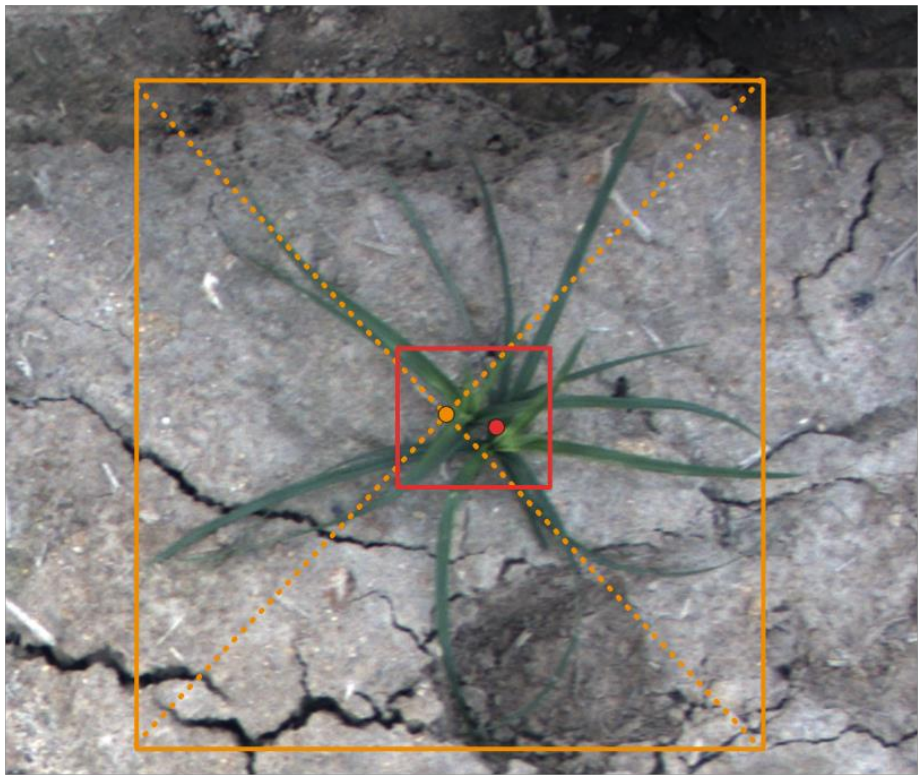} 
        \caption{}
        \label{fig:subfig-a1}
    \end{subfigure}
    \hfill
    \begin{subfigure}[b]{0.48\linewidth}
        \centering
        \includegraphics[height=4cm,width=\textwidth]{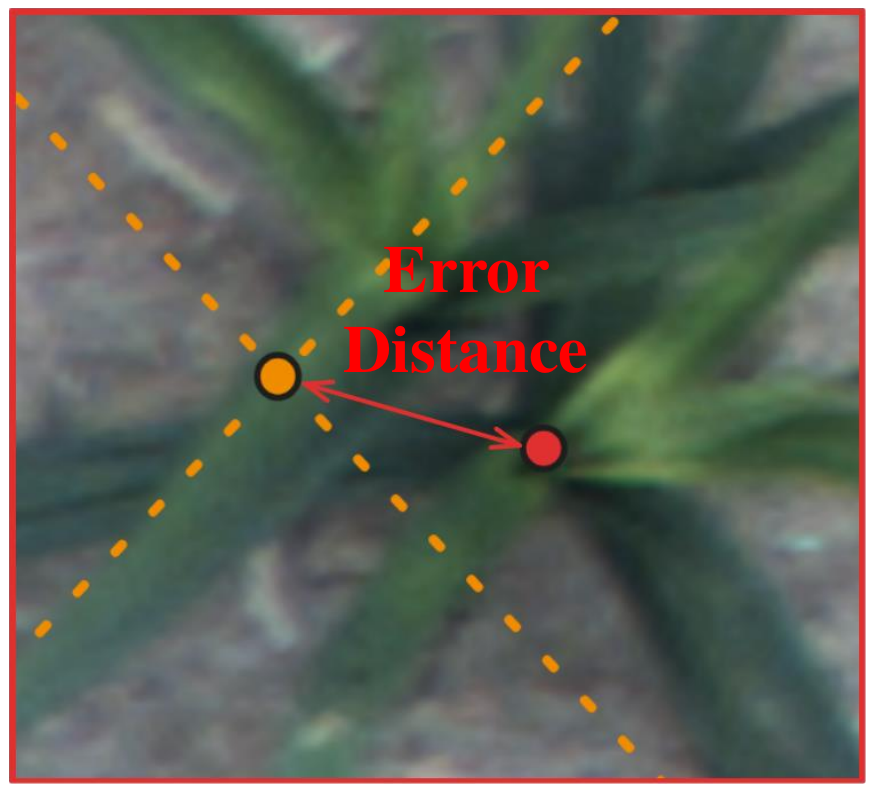} 
        \caption{}
        \label{fig:subfig-a2}
    \end{subfigure}

    \begin{subfigure}[b]{1\linewidth} 
        \centering
        \includegraphics[width=\textwidth]{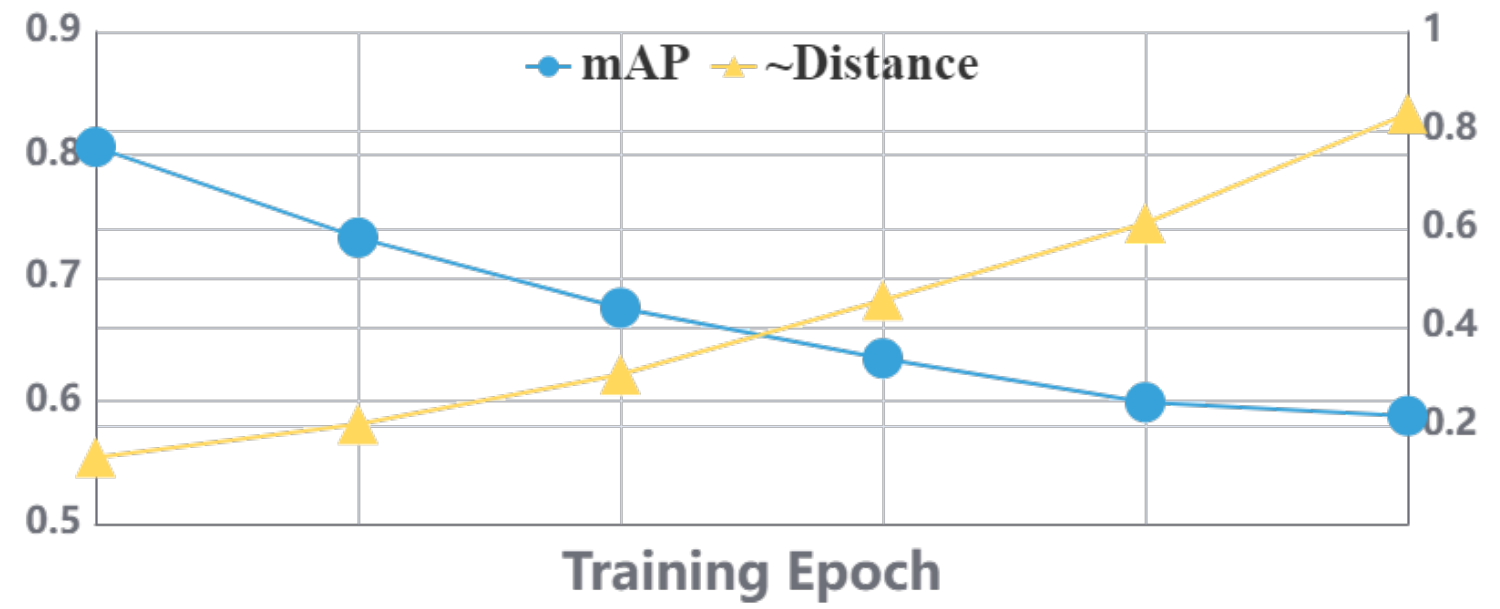}
        \caption{}
        \label{fig:subfig-b}
    \end{subfigure}
    
    \caption{(a) Weed detection with YOLOv7: Yellow dashed lines mark bounding box diagonals, indicating the geometric center, while the blue dot shows the ground-truth weed stem location. (b) Close-up of the red-boxed region, highlighting misalignment between the bounding box center and the ground-truth point. (c) In laser weeding, better weed detection performance (lower mAP) does not always mean better weed stem detection (lower Euclidean Distance).}
    \label{fig:map_vs_mnd}
\end{figure}


\section{Introduction}

Sustainable agricultural management is essential for addressing global hunger and achieving the United Nations' ``Zero Hunger'' goal and the principle of ``Leaving No One Behind'' (LNOB)~\cite{UnitedNations2023b}. Effective weed control plays a non-trivial role in maintaining food security, as weeds compete with crops for critical resources such as water, nutrients, and sunlight, which directly affects crop yield and quality. 

Current weed control methods are generally categorized into chemical weeding and mechanical weeding. Chemical methods usually use toxic substances to inhibit or destroy weeds at various growth stages, including those applied before or after weeds emerge. Although effective against different weed types, these methods can negatively influence the crop quality and inevitably cause chemical soil degradation, resulting in environmental pollution~\cite{zhang1996,Zhang1996b,tanveer2003}. Instead, mechanical weed control methods use machines such as mowers~\cite{pirchio2018,sportelli2020autonomous,aamlid2021robo}. Mowers often miss small weeds and can only cut the part of weed above the surface. Thus, mowers are not effective when dealing with deeply-rooted and perennial ones, leading to frequent maintenance. A desired weed control approach should take both efficiency and environmental friendliness into consideration. Laser weeding~\cite{weedlazer,LaserW2023} offers a promising alternative with aforementioned properties. It leverages high-energy and high-temperature laser beam to target and cut weeds at the stem, effectively killing the weeds. Additionally, laser weeding can be environmentally friendly if it is powered by clean energy.  

Fueled by recent advances in deep learning, weed recognition has been well studied theoretically~\cite{wu2021review,hu2024deep}.
Considering that laser weeding has the advantages in weeding efficiency and energy consumption, intelligent laser weeding seems to be a promising path for society. On the contrary, efficiency and energy become two critical issues for intelligent laser weeders. To conserve high-energy beams, the diameter of laser transmitters is much smaller in size compared with the leaves. Simply detecting or segmenting the weed is not a effective signal to transmit the laser beam as cutting leaves can not eradicate the weed. To maintain high efficiency in terms of energy usage, the laser beam shall be aimed directly at the bottom of weed stem. Meanwhile, as the laser can cause irreversible damage on crops, the detection algorithms require low false positive rate. So far, though there have been a few start-ups trying to address this challenge~\cite{weedlazer,LaserW2023}, accurately locating the weed stem remains a challenge~\cite{zhang2023}.
A visual illustration is presented in Fig.~\ref{fig:map_vs_mnd}. Based on geometric principle and agricultural knowledge, weed stems are intuitively expected to align with the geometric center of the detected bounding box in a vertical view. Though the predicted bounding boxes can achieve high mAP, the predicted location of weed stem is far from the ground truth location (Fig.~\ref{fig:map_vs_mnd}(b)). Moreover, traditional weed detection methods typically use mAP as the evaluation metric, which may not be suitable for the task of laser weeding. As shown in Fig.~\ref{fig:map_vs_mnd}(c), a method with a high mAP score may still exhibit poor root localization performance, but improving distance accuracy is crucial as it directly benefits the effectiveness of laser weeding.

To tackle the aforementioned challenges, we propose a pipeline that integrates crop and weed detection with weed stem localization into a unified end-to-end system. Specifically, we introduce an additional root coordinate regression branch within the object detection framework. The proposed system can process a sequence of images or a real-time video stream, detecting plant bounding boxes and simultaneously pinpointing weed stems for laser transmission, thereby ensuring effective weed control without damaging crops. This pipeline is simple yet robust, and can be easily implemented in object detectors.
To train and validate the proposed system in real-world conditions, and to empirically understand the task of weed recognition under the setup of laser weeding, we collect and curate the Weed Stem Detection (WSD) dataset, consisting of 7,161 high-resolution images with 11,151 annotated instances. This dataset includes bounding boxes for three crops and weeds, as well as the coordinates of weed stem. The main contributions of this work are summarized below.

\begin{itemize}
\item We provide a high-quality weed stem dataset with human annotations and the first empirical study on weed recognition for practical laser weeding, addressing a significant academic gap in laser weeding.
\item We propose an end-to-end deep learning pipeline that integrates crop and weed detection with weed stem localization and can be extended to semi-supervised learning, which can further leverage unlabeled data.
\item We experimentally demonstrate that our method is more efficient than previous detection-based methods in laser weeding by improving the weeding accuracy by 6.7\% and reducing the energy cost by 32.3\%.
\end{itemize}

\section{Related Work}
\subsection{Weed Datasets}

Existing weed datasets primarily address weed recognition or detection~\cite{hasan2021survey, hu2024deep}. We summarize the key public datasets with human annotations in Tab.~\ref{tab:weed_datasets}. DeepWeeds~\cite{olsen2019deepweeds} includes 17,509 images of eight Australian weeds but lacks crop data, limiting its practical weeding applications. The Weed-Corn/Lettuce/Radish dataset~\cite{jiang2020cnn} contains 7,200 images with four species (three crops and one weed), while the Food Crop and Weed Dataset~\cite{sudars2020dataset} includes 1,118 images of seven species (six crops and one weed). CottonWeedID15~\cite{chen2022performance} consists of 5,187 weed images from cotton fields with image-level annotations only. CottonWeedDet12~\cite{lu2023cottonweeddet12} and CottonWeedDet3~\cite{rahman2023performance} add bounding box annotations. The largest dataset, CropAndWeed~\cite{steininger2023cropandweed}, has coarse machine-generated labels. However, precise weed stem localization is essential for laser weeding, requiring high-quality annotations in terms of localization. Although some laser-weeding datasets exist~\cite{zhang2024laser}, they focus on classification, detection, and segmentation rather than precise stem localization. To the best of our knowledge, WSD is the first dataset with human annotations for both crop and weed detection, as well as weed stem localization.

\subsection{Weed Recognition}
Existing studies on weed recognition can be broadly categorized into four tasks: weed classification, weed object detection, weed object segmentation, and weed instance segmentation~\cite{hu2024deep}. Weed classification focuses on identifying weeds at the image level, determining whether an image contains non-crop plants. For instance, SVM classifiers have achieved about 95\% accuracy in relatively simple environments~\cite{zhang2022segmentation}, and by combining VGG with 
SVM~\cite{tao2022hybrid}, a 99\% accuracy rate has been reached in distinguishing between weeds and grapevines. Weed object detection extends beyond classification by providing bounding boxes to locate weeds within images.~\cite{parra2020edge,nasiri2022deep} Various models have been successfully applied to this task, including DetectNet~\cite{yu2019weed}, Faster R-CNN~\cite{veeranampalayam2020comparison}, and YOLOv3~\cite{sharpe2020vegetation}, all showing promising results. Weed object segmentation and instance segmentation focus on pixel-level recognition~\cite{jeon2011robust, long2015fully, you2020dnn}, offering more detailed analysis. For example, VGG-UNet has been used to segment sugar beets and weeds~\cite{fawakherji2019uav}. 
However, none of these methods can localize the weed stem, a crucial aspect of effective weed management. To address this gap, our work introduces an end-to-end framework that simultaneously detects crops and weeds while localizing the weed stem.

\begin{table*}[t!]
\centering
\setlength{\tabcolsep}{2.5mm}
\renewcommand{\arraystretch}{1.5}
\begin{tabular}{ccccccc}
\hline
Dataset & Stem & BBox & \# Img & \# Species & \# Inst & Res\\
\hline
DeepWeeds~\shortcite{olsen2019deepweeds} & \usym{2715} & \usym{2715} & 17509 & 1 & - & 256 $\times$ 256 \\ 
Weed-Corn/Lettuce/Radish~\shortcite{jiang2020cnn} & \usym{2715} & \usym{2715} & \phantom{0}7200 & 4 & - & 800 $\times$ 600  \\
Crops and Weed Dataset~\shortcite{sudars2020dataset} & \usym{2715} & \usym{2713} & \phantom{0}1118 & 7 & - & [480, 1000] $\times$ [384, 1280] \\
CottonWeedID15~\shortcite{chen2022performance} & \usym{2715} &  \usym{2715} & \phantom{0}5187 & 1 & - & 512 $\times$ 512 \\
CottonWeedDet3~\shortcite{rahman2023performance} & \usym{2715} & \usym{2713} & \phantom{0}\phantom{0}848 & 1 & \phantom{0}1.8 $\pm$ 1.5 & 4442 $\times$ 4335 \\
CottonWeedDet12~\shortcite{lu2023cottonweeddet12} & \usym{2715} & \usym{2713} & \phantom{0}5648 & 1 & \phantom{0}1.7 $\pm$ 1.4 & 3024 $\times$ 4032 \\
\hline
WSD (Ours) & \usym{2713} & \usym{2713} & \phantom{0}7161 & 4 & 12.5 $\pm$ 7.5  & 2048 $\times$ 2048 \\
\hline
\end{tabular}
\caption{Comparison between WSD dataset and existing weed datasets with human annotations. ``Stem'' indicates whether stem annotations are provided. ``\#Img'' denotes the number of images. ``\# Species'' denotes the number of species. ``\# Inst'' denotes the average number of annotations per image, along with the standard deviation. ``Res'' denotes the image resolution.}
\label{tab:weed_datasets}
\end{table*}

\begin{figure}[ht]
\centering
\begin{minipage}[t]{0.45\linewidth}
\includegraphics[width=1\linewidth]{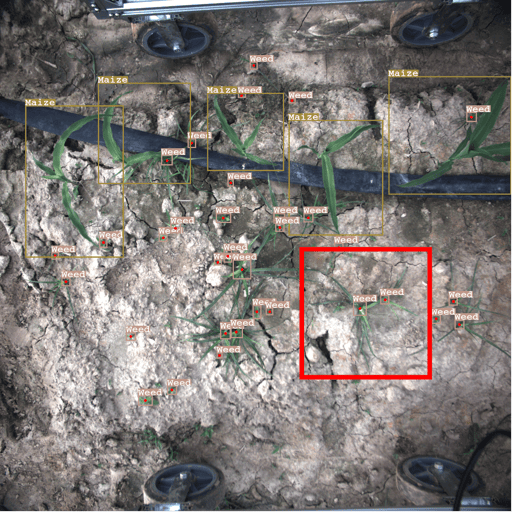}
\includegraphics[width=1\linewidth]{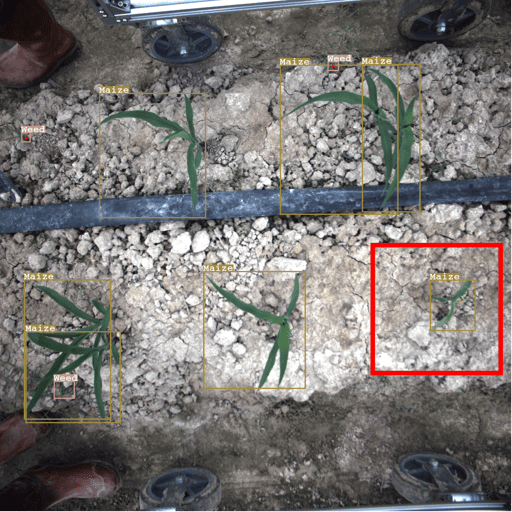}
\includegraphics[width=1\linewidth]{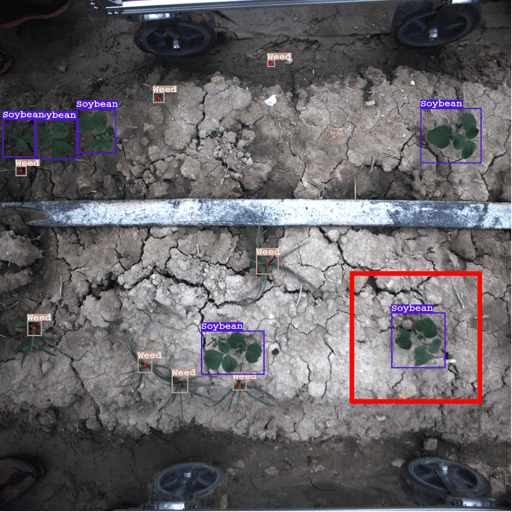}
\includegraphics[width=1\linewidth]{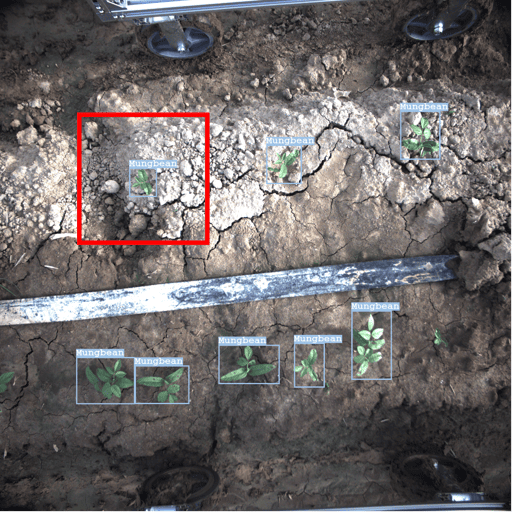}
\centerline{(a) Raw Image}
\end{minipage}
\begin{minipage}[t]{0.45\linewidth}
\includegraphics[width=1\linewidth]{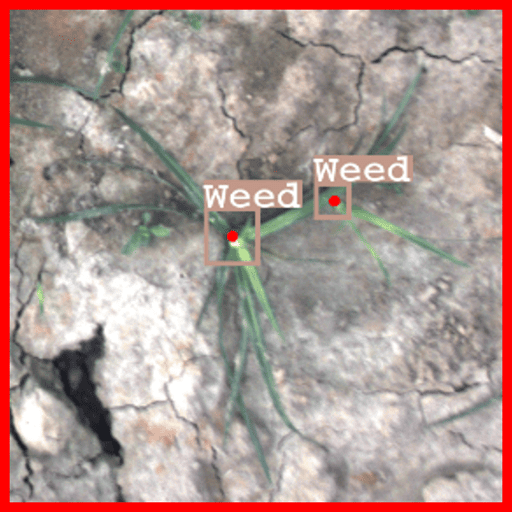}
\includegraphics[width=1\linewidth]{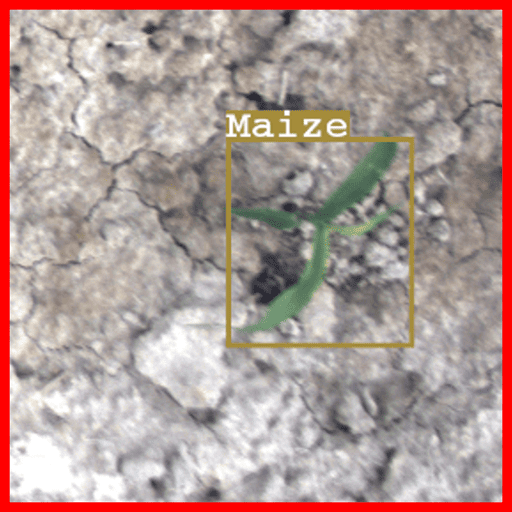}
\includegraphics[width=1\linewidth]{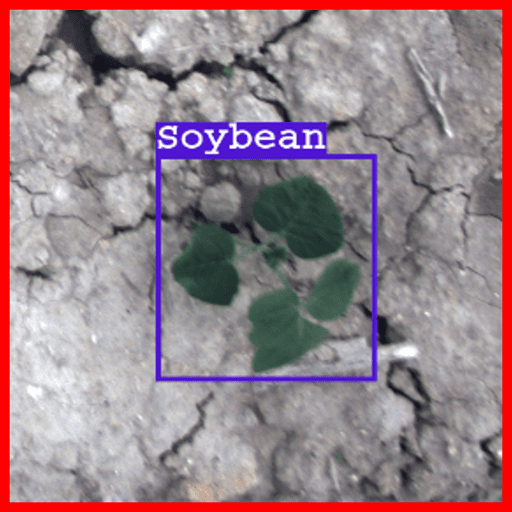}
\includegraphics[width=1\linewidth]{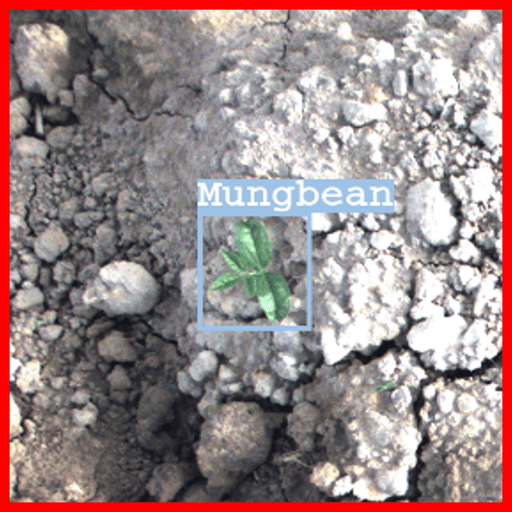}
\centerline{(b) Zoomed-in View}
\end{minipage}
\caption{Image samples show raw images (left) and 16x zoomed sections (right), highlighting four different species.}
\label{fig:example}
\end{figure}

\begin{figure}[ht]
\centering
\includegraphics[width=\linewidth]{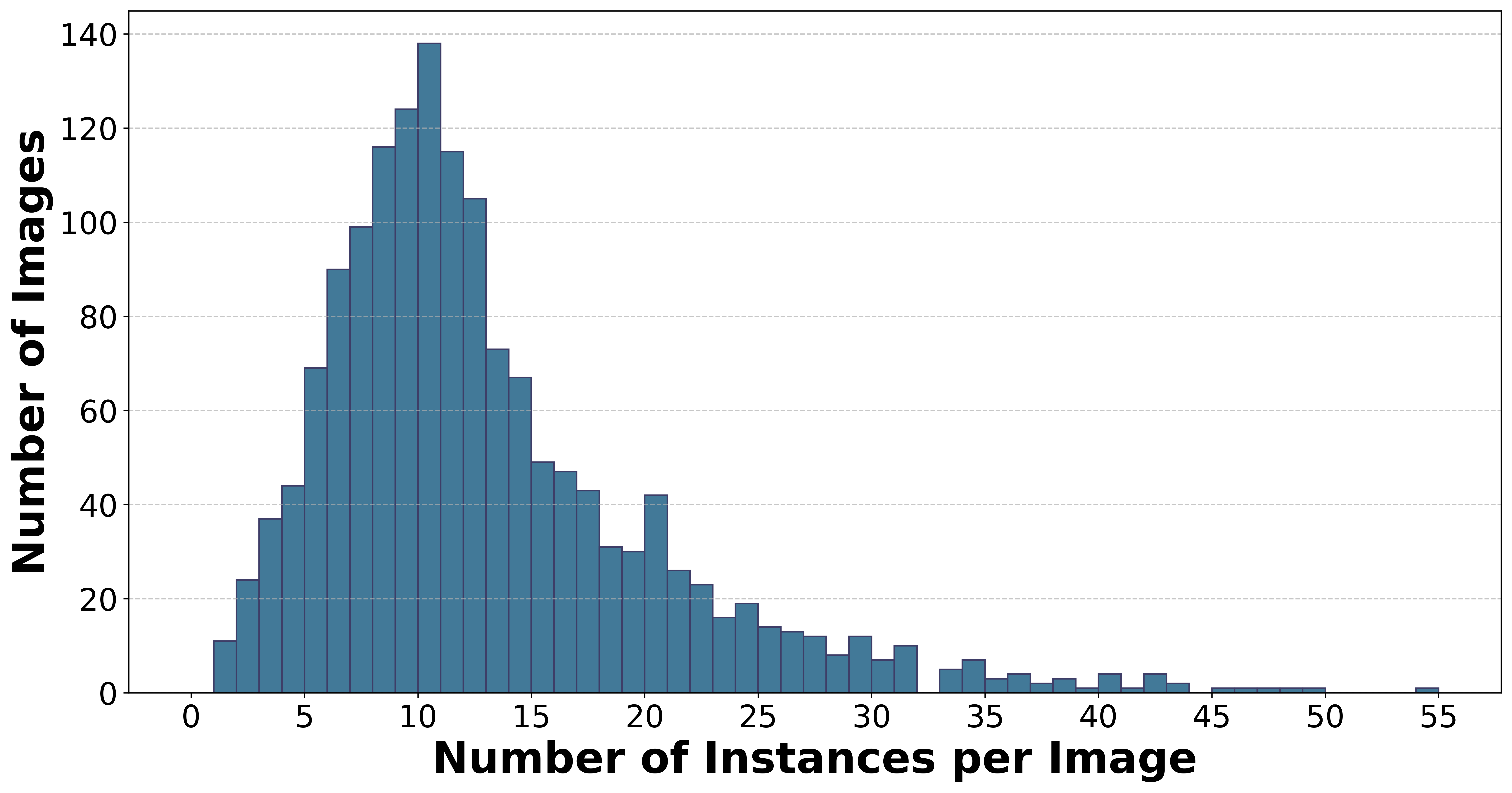}
\caption{Distribution of instance annotations per image: The X-axis shows the number of instances (including both bounding box and point annotations) per image, while the Y-axis indicates the corresponding image count.} 
\label{dist_weed}
\end{figure}


\section{Weed Stem Detection Dataset}

\subsection{Data Collection}
The standard RGB images are collected by a custom-built autonomous vehicle equipped with Teledyne FLIR BFS-U3-123S6C-C, a high-resolution imagery sensor. Each image has a resolution of $2048 \times 2048$. 
The sensor is embedded in the autonomous vehicle, making the sensor at a relatively fixed height above the surface, which is one meter for the prototype vehicle. The images are captured in three different experimental fields planted with three different crops: maize, soybean, and mungbean, respectively. 
We intentionally planted weed seeds in the field at staggered intervals, resulting in weeds at various growth stages. All crops, however, are at the seedling stage, 30 days after sowing--an important period for weeding.

\subsection{Data Annotation}
We deployed LabelImg\footnote{\url{https://pypi.org/project/labelImg/}}, a graphical image annotation tool. The human annotators can label object bounding boxes in images with LabelImg, which saves the annotation details in XML files. Three professional agronomists with advanced graduate degrees and field experience were hired to complete the annotation. The annotation process was completed in two steps. First, the bounding boxes of crop and weed were annotated. Then, weed stem locations were annotated in a point coordinate fashion. All final annotations were verified by all three agronomists to achieve consensus. Any discrepancies among the human annotators were resolved through re-annotation to ensure reliability. Fig.~\ref{fig:example} shows four annotated images with zoomed-in visualization. 

\subsection{Dataset Statistics}
There are 7,161 images in total, with 1,556 annotated and 5,605 unannotated images. The inclusion of unannotated images allows the dataset to be extended for semi-supervised learning scenarios. The distribution of instance annotations per image is illustrated in Fig.~\ref{dist_weed}. It is worth mentioning that the annotation is time-consuming. On average, it takes approximately 135 seconds to label a weed instance. The statistics of WSD are summarized in Tab.~\ref{tab:weed_info}.

\begin{table}[ht]
    \centering
    \setlength{\tabcolsep}{1.5mm}
    \renewcommand{\arraystretch}{1.5}
    \begin{tabular}{ccccc}
         \hline
         Class &  \# Instances &  Share  & \# Images &  Time (s)\\ 
         \hline
         Weed & 11151 &  57.3\% & 1379 & 135 $\pm$ 45 \\
         Maize & \phantom{0}5808 & 29.9\% & 1175 & 105 $\pm$ 15 \\
         Soybean & \phantom{0}1624 & \phantom{0}8.4\% & \phantom{0}268 & \phantom{0}60 $\pm$ 10 \\
         Mungbean & \phantom{0}\phantom{0}848 & \phantom{0}4.4\% & \phantom{0}100 & \phantom{0}75 $\pm$ 15 \\
         \hline
    \end{tabular}
    \caption{Statistics of Weed Stem Detection Dataset. ``Instances'' indicates the number of bounding boxes with additional point annotations. ``Share'' represents the percentage of instances in this category. ``Images'' is the number of images containing this category. ``Time'' refers to the average $\pm$ standard deviation of annotation time in seconds by professional agronomists.}
    \label{tab:weed_info}
\end{table}


\section{Method}
\begin{figure*}[ht]
  \centering
    \includegraphics[width=1\linewidth]{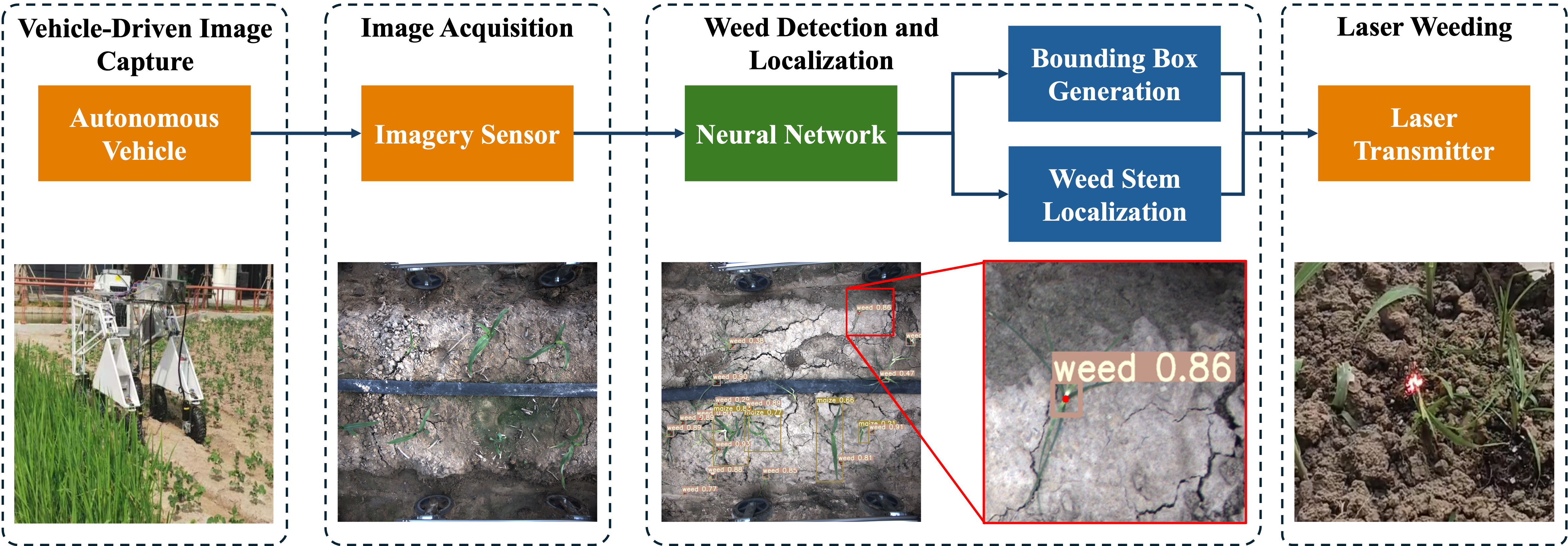}
    \caption{The pipeline of intelligent laser weeding. The autonomous vehicle captures the image (or video frame). The proposed neural network infers the class and location of each crop and weed.  Upon identifying a weed, the model also outputs the stem location of the weed, followed by a laser beam cut.}
    \label{fig:pipeline}
\end{figure*}

The proposed laser weeding pipeline is depicted in Fig.~\ref{fig:pipeline}. The autonomous vehicle first captures images that include both crops and weeds. These images are then processed by a neural network to detect and localize weed stems. Upon detection, a laser beam is emitted to cut the weed stems. This section details the integration of stem regression within a pre-trained object detection neural network and the subsequent enhancement of its performance using semi-supervised learning.

\subsection{Weed Stem Regression}
To accurately localize weed stems, we augment the pre-trained object detection neural network $NN(\cdot)$ with an additional stem coordinate regression head, formulated as:
\begin{gather}
E_i = NN(I_i) \\
\hat{y_i} = Conv(E_i)
\end{gather}
where $I_i$ represents the $i$-th input image, $E_i$ denotes the extracted image embedding, $\hat{y_i}$ is the predicted stem coordinate, and $Conv(\cdot)$ is the regression head. The regression loss $L_{reg}$ is computed as:
\begin{equation}
L_{reg} = \frac{1}{n} \sum_{i=1}^{n}MSE(y_{i}, \hat{y}_{i}) 
\end{equation}
$MSE(\cdot)$ represents the Euclidean Distance calculation, $y_{i}$ is the ground truth weed stem coordinate, and only weed coordinates are used in calculating the regression loss. To jointly optimize bounding box detection and weed stem regression, we combine the regression loss $L_{reg}$ with the classification loss $L_{cls}$ and the bounding box detection loss $L_{bbox}$ as follows:
\begin{gather}
L = \alpha \cdot L_{cls} + \beta \cdot L_{bbox} + \gamma \cdot L_{reg}
\end{gather}
where $\alpha$, $\beta$, and $\gamma$ are hyper-parameters that balance the contributions of the different losses.

\subsection{Extension to Leverage Unlabelled Images}
\begin{figure}[ht]
    \centering
    \includegraphics[width=\linewidth]{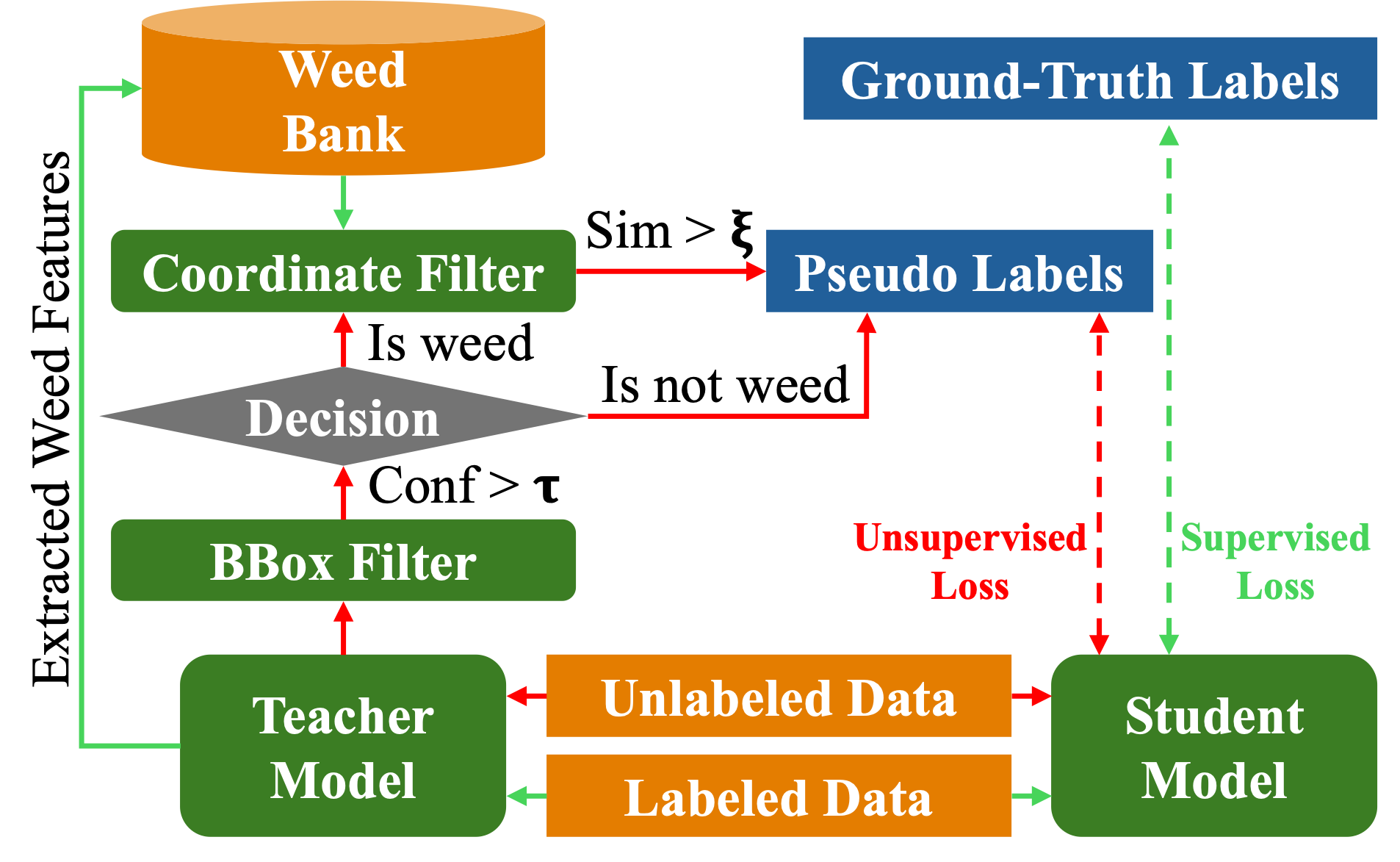}
    \caption{Overview of semi-supervised learning process. Pseudo labels are first generated for unlabeled data using a teacher model, followed by training a student model with both labeled and pseudo-labeled data. ``Conf'' represents the confidence score for classification, and ``Sim'' denotes the cosine similarity between extracted ground-truth weed embeddings and predicted weed embeddings, used to filter out low-quality weed localization. $\tau$ and $\xi$ are hyper-parameters.}
    \label{fig:semi-supervised}
\end{figure}

Labeling images in real-world scenarios is labor-intensive. To reduce annotation costs and simultaneously leverage unlabeled images to enhance model performance, we employ a teacher-student framework for semi-supervised learning~\cite{kingma2014semi,zhai2019s4l,berthelot2019mixmatch,xu2021end}. As illustrated in Fig.~\ref{fig:semi-supervised}, pseudo labels for the unlabeled data $\mathcal{D}_u$ are generated using a teacher model. The student model is then trained on both the labeled data $\mathcal{D}_l$ and the pseudo-labeled data $\mathcal{D}_p$.

\subsubsection{Pseudo Label Generation}
To effectively utilize the abundant unlabeled images, we first fine-tune a teacher model $Teacher(\cdot)$ based on the pre-trained neural network with the combined loss $L$. The fine-tuned teacher model classifies unlabeled images, assigning pseudo labels to predictions with confidence higher than the threshold $\tau$. Confidence $ConfScore$ is calculated as:
\begin{gather}
ConfScore = Max(Softmax(Teacher(E_u)))
\end{gather}
where $E_u$ denotes unlabeled image embeddings (subscripts are omitted for simplicity). Since precise weed coordinate prediction is crucial, we use ground-truth weed image embeddings as anchors to filter out low-quality predictions. Specifically, we extract weed embeddings $E_l^w$ from labeled data and store them in a weed bank. The cosine similarity between predicted weed embeddings $E_u^w$ and all pre-extracted weed embeddings $E_l^w$ is then calculated:
\begin{gather}
SimScore = CosineSimilarity(E_l^w, E_u^w)
\end{gather}
Predictions with $SimScore$ higher than the threshold $\xi$ are assigned as pseudo labels. Finally, the student model is trained on both labeled data $\mathcal{D}_l$ and pseudo-labeled data $\mathcal{D}_p$. Notably, weak and strong augmentations are applied to each unlabeled image: the weakly augmented images are fed into the student network, while the strongly augmented images are processed by the teacher network. Weak augmentations include adjustments to brightness and contrast, while strong augmentations additionally involve cropping and flipping.

\begin{table}[ht]
\centering
\setlength{\tabcolsep}{1.5mm}
\renewcommand{\arraystretch}{1.5}
\begin{tabular}{ccccc}
\hline
Model & Dist$\downarrow$ & FP & Accuracy & Cost \\
\hline
Detection & 2.9770 & 0 & 75.37\% & 1.55 \\
Ours & 2.4838 & 0 & 80.42\% & 1.05 \\
\hline
\end{tabular}
\caption{Effect of stem regression. ``FP'' denotes false positive rate, \emph{i.e.} the crops are identified as the weeds. ``Accuracy'' denotes the weeding accuracy. ``Cost'' denotes the energy cost where the unit cost denotes the hypothetical minimum energy cost with ground truth-level prediction, \emph{i.e} one weed only requires one shot.}
\label{tab:detection_vs_root_regression}
\end{table}

\begin{figure*}[!ht]
    \centering
    
    \raisebox{0.3\height}{\rotatebox{90}{Detection}\hspace{1.5mm}}%
    \begin{subfigure}{0.16\linewidth}
        \includegraphics[width=\linewidth]{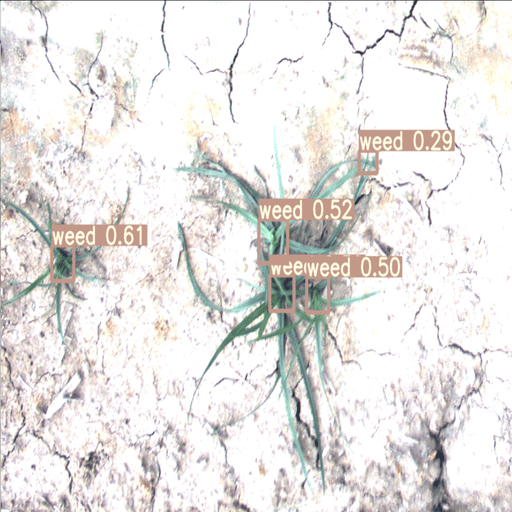}
    \end{subfigure}%
    \begin{subfigure}{0.16\linewidth}
        \includegraphics[width=\linewidth]{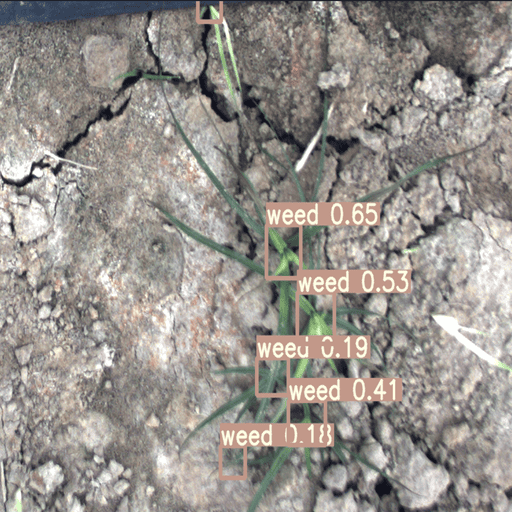}
    \end{subfigure}%
    \begin{subfigure}{0.16\linewidth}
        \includegraphics[width=\linewidth]{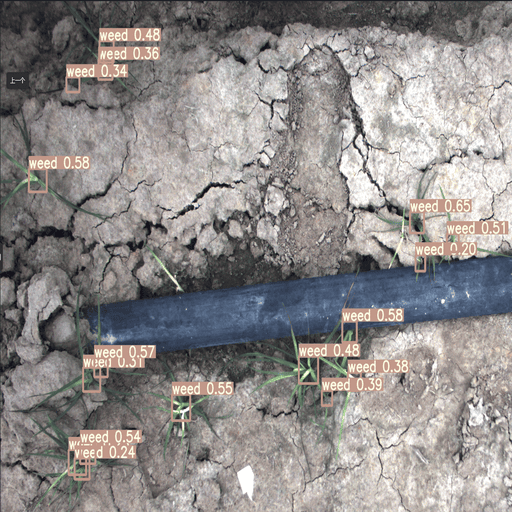}
    \end{subfigure}%
    \begin{subfigure}{0.16\linewidth}
        \includegraphics[width=\linewidth]{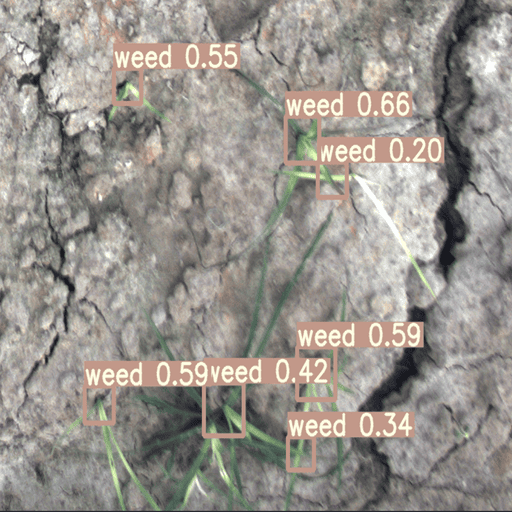}
    \end{subfigure}%
    \begin{subfigure}{0.16\linewidth}
        \includegraphics[width=\linewidth]{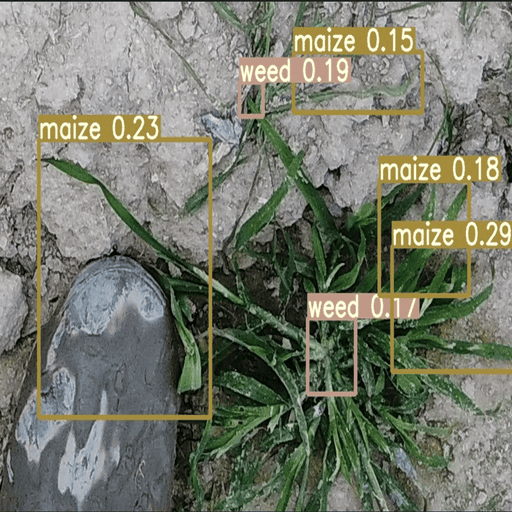}
    \end{subfigure}%
    \begin{subfigure}{0.16\linewidth}
        \includegraphics[width=\linewidth]{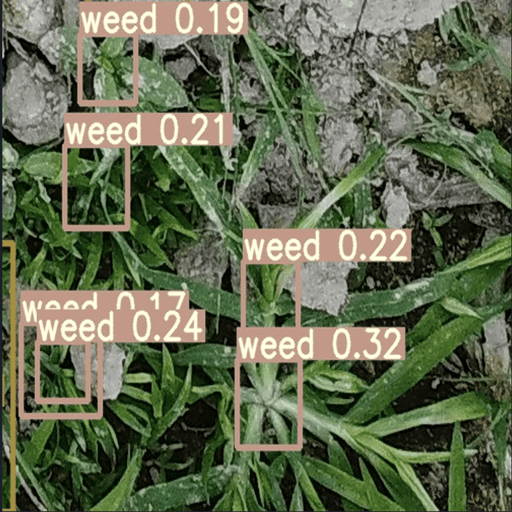}
    \end{subfigure}

    \raisebox{1.2\height}{\rotatebox{90}{Ours}\hspace{1.5mm}}%
    \begin{subfigure}{0.16\linewidth}
        \includegraphics[width=\linewidth]{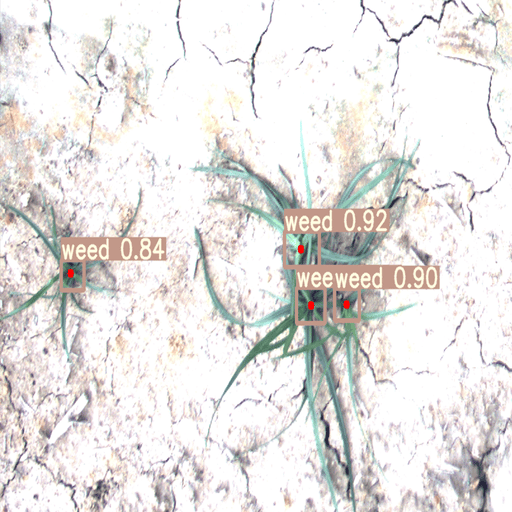}
    \end{subfigure}%
    \begin{subfigure}{0.16\linewidth}
        \includegraphics[width=\linewidth]{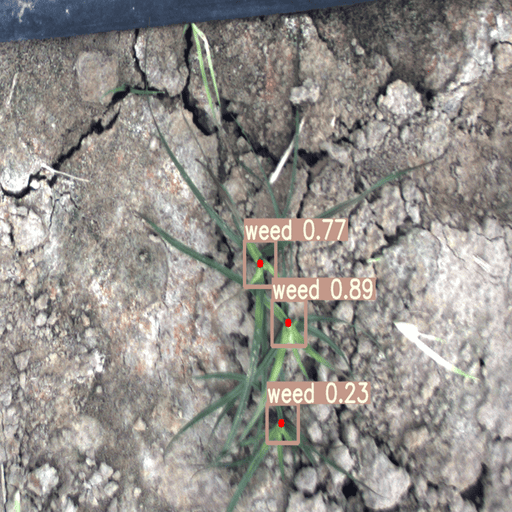}
    \end{subfigure}%
    \begin{subfigure}{0.16\linewidth}
        \includegraphics[width=\linewidth]{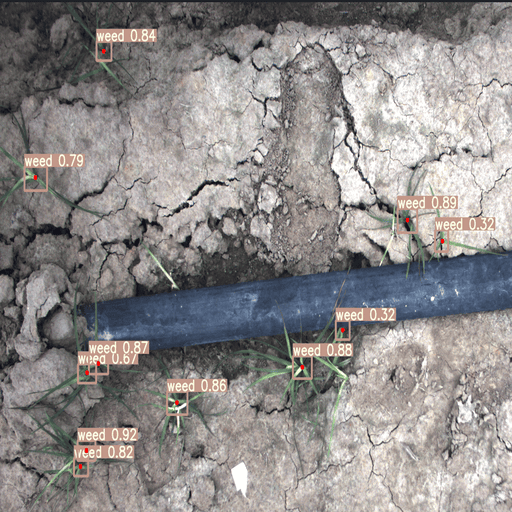}
    \end{subfigure}%
    \begin{subfigure}{0.16\linewidth}
        \includegraphics[width=\linewidth]{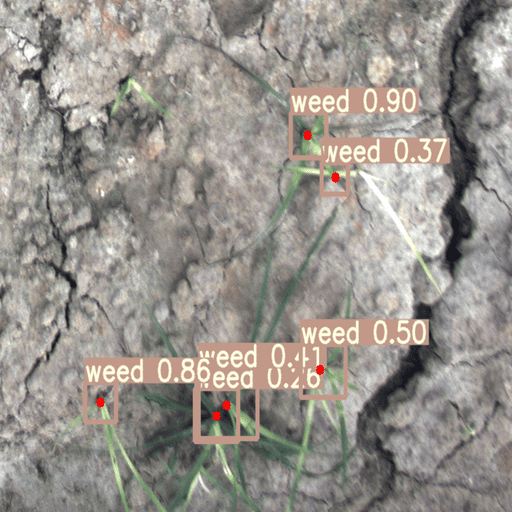}
    \end{subfigure}%
    \begin{subfigure}{0.16\linewidth}
        \includegraphics[width=\linewidth]{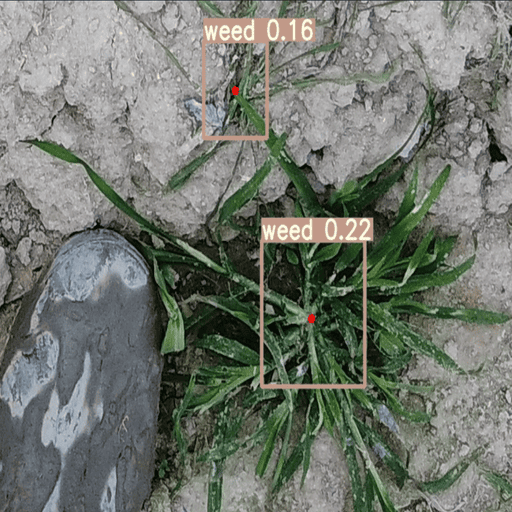}
    \end{subfigure}%
    \begin{subfigure}{0.16\linewidth}
        \includegraphics[width=\linewidth]{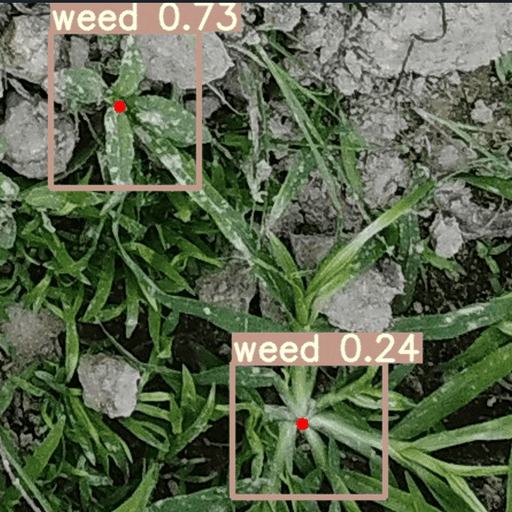}
    \end{subfigure}

    \raisebox{2\height}{\rotatebox{90}{GT}\hspace{1.5mm}}%
    \begin{subfigure}{0.16\linewidth}
        \includegraphics[width=\linewidth]{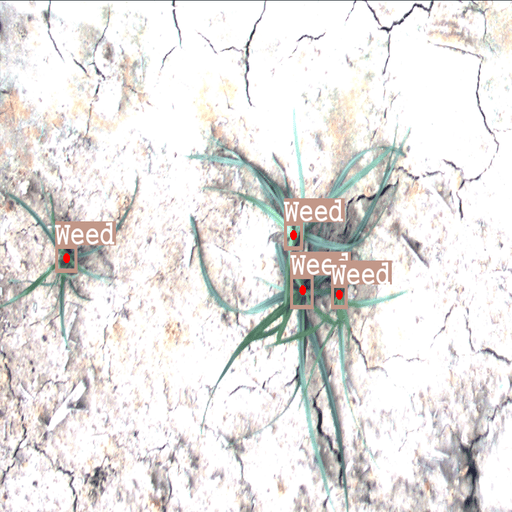}
    \end{subfigure}%
    \begin{subfigure}{0.16\linewidth}
        \includegraphics[width=\linewidth]{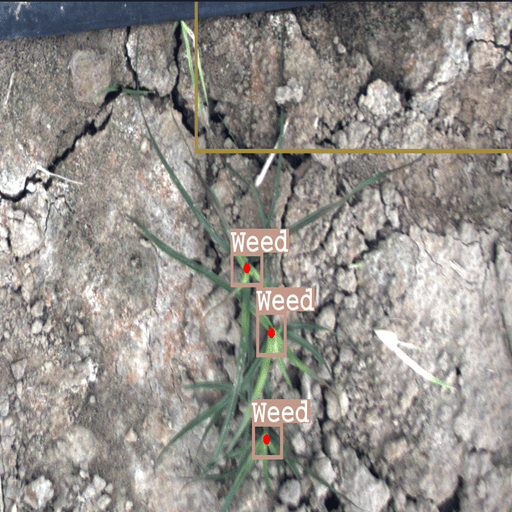}
    \end{subfigure}%
    \begin{subfigure}{0.16\linewidth}
        \includegraphics[width=\linewidth]{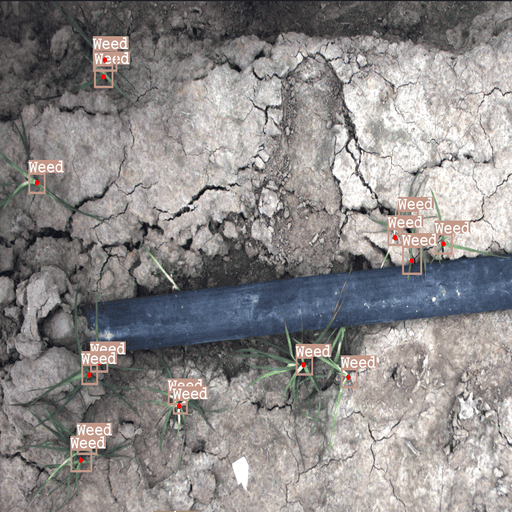}
    \end{subfigure}%
    \begin{subfigure}{0.16\linewidth}
        \includegraphics[width=\linewidth]{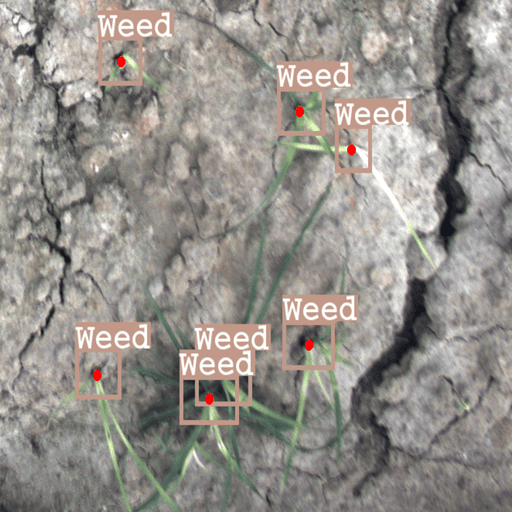}
    \end{subfigure}%
    \begin{subfigure}{0.16\linewidth}
        \includegraphics[width=\linewidth]{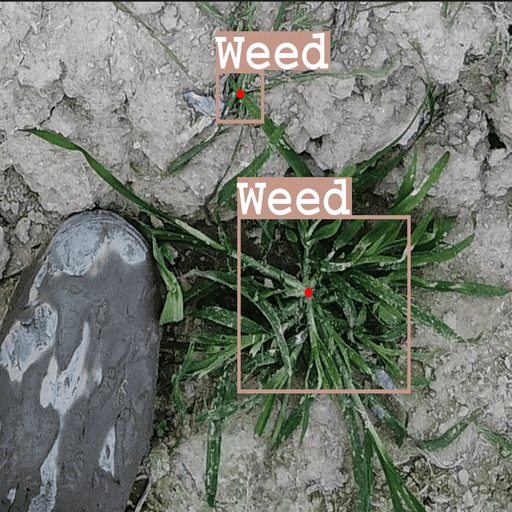}
    \end{subfigure}%
    \begin{subfigure}{0.16\linewidth}
        \includegraphics[width=\linewidth]{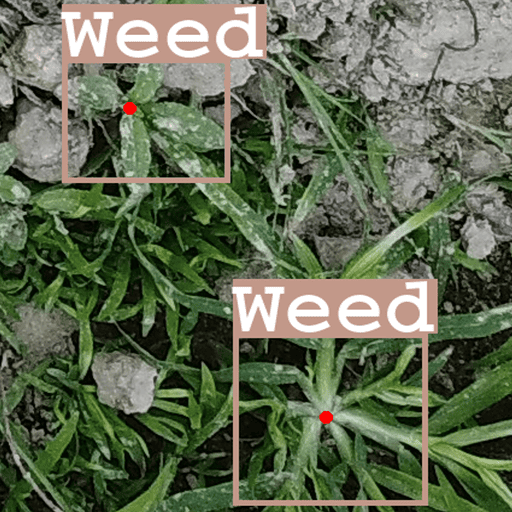}
    \end{subfigure}
    
    \caption{Qualitative comparison of weed detection results between the vanilla detection method and our method. The results of our method (second row) shows higher true positive rate than the vanilla detection method (first row), especially when a weed spread and looks like multiple weeds, thus conserving the energy.}
    \label{fig:inference_results}
\end{figure*}

\section{Experiments}
\subsection{Experimental Setup}
\subsubsection{Implementation}

We conduct empirical evaluations of our method on the WSD dataset, using $80\%$ of the data for training, $10\%$ for validation, and $10\%$ for testing. Given the hardware limitations of the autonomous vehicle, we prioritize lightweight deployment and real-time inference by selecting YOLOv7~\cite{wang2023} as the baseline detection model due to its superior performance among existing object detection methods~\cite{dang2023yoloweeds, rahman2023performance}. All models are trained using an SGD optimizer with a learning rate of 1e-3 for 300 epochs. We set the loss weights $\alpha$, $\beta$, and $\gamma$ to 0.2, 0.3, and 0.5, respectively. For semi-supervised learning, we adopt a dynamic updating mechanism with EMA, where the smoothing factor is set to 0.9, the confidence threshold is set to 0.5 and the cosine similarity threshold to 0.4. To be mentioned, all hyper parameters are determined on the basis of empirical results. The experiments are conducted on a single NVIDIA Tesla A100 80G GPU.
 
\subsubsection{Evaluation Metric}
While mAP is a well-established metric that aggregates precision and recall to provide an overall measure of detection performance, it primarily evaluates the presence and classification of objects rather than their precise positioning. As shown in Fig.~\ref{fig:map_vs_mnd}, mAP is not a robust measure for the task of interest.
Instead, we evaluate using Euclidean Distance (Dist), where a lower value indicates higher accuracy in weed stem localization.

\subsection{Results}
\subsubsection{Effect of Stem Regression}

\begin{figure}[ht]
    \centering
    \begin{subfigure}{0.48\linewidth}
        \centering
        \includegraphics[width=\linewidth]{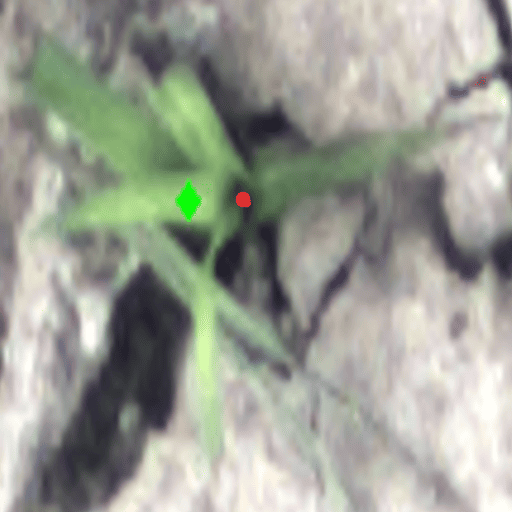}
    \end{subfigure}
    \begin{subfigure}{0.48\linewidth}
        \centering
        \includegraphics[width=\linewidth]{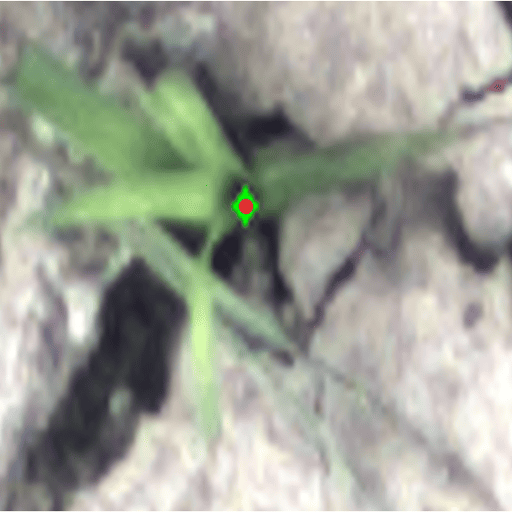}  
    \end{subfigure}

    \begin{subfigure}{0.48\linewidth}
        \centering
        \includegraphics[width=\linewidth]{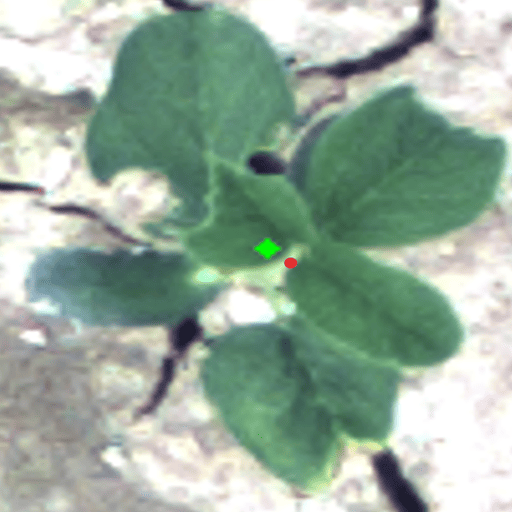}
    \end{subfigure}
    \begin{subfigure}{0.48\linewidth}
        \centering
        \includegraphics[width=\linewidth]{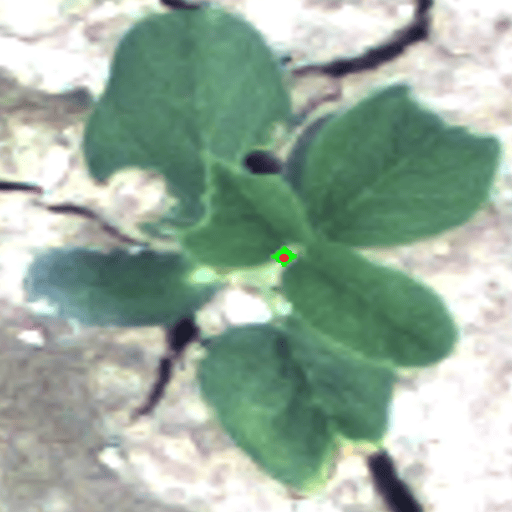}
    \end{subfigure}

    \begin{subfigure}{0.48\linewidth}
        \centering
        \includegraphics[width=\linewidth]{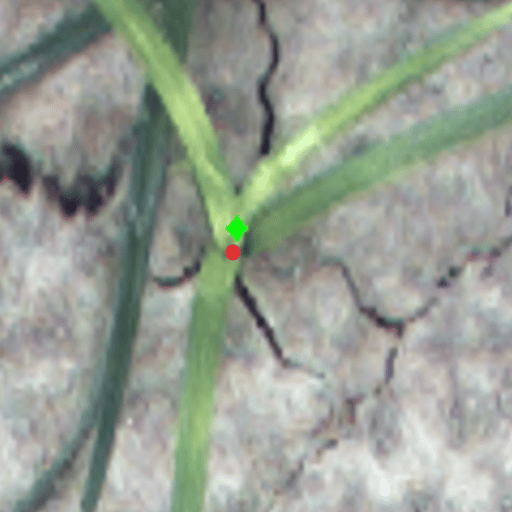}
        \caption{Detection}
    \end{subfigure}
    \begin{subfigure}{0.48\linewidth}
        \centering
        \includegraphics[width=\linewidth]{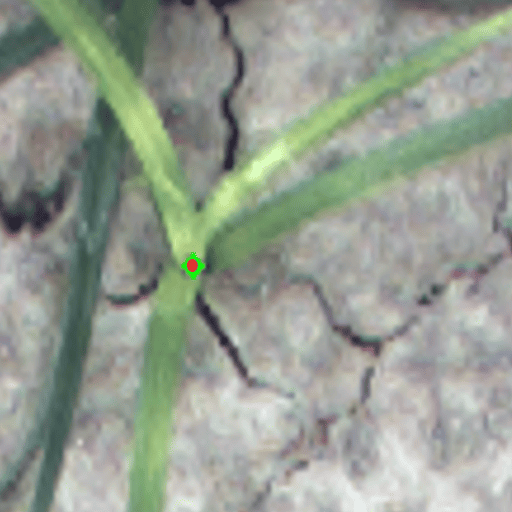}
        \caption{Ours}
    \end{subfigure}
    \caption{Qualitative comparison of stem localization results between the vanilla detection method (a) and our method (b). Green and red points denote the prediction and the ground truth, respectively. Our method consistently outperforms the vanilla detection method.}
    \label{fig:comp_with_gt_center}
\end{figure}

In this empirical study, we first validate the effectiveness of stem regression, which significantly reduces the Dist value (Tab.~\ref{tab:detection_vs_root_regression}). We then design a real-world simulated experiment to assess the efficiency of the proposed system. Under identical experimental conditions, the vanilla detection method and our method process the same number of weeds independently. For the vanilla method, the geometric center of the predicted bounding box is used as the stem coordinate. We evaluate two aspects: weeding accuracy (percentage of weeds eradicated) and energy consumption (laser shots). As shown in Tab.~\ref{tab:detection_vs_root_regression}, our method reduces energy cost by up to 32.3\% while improving accuracy by 6.7\%, with no misidentification of crops as weeds. This suggests a clear visual difference between crops and weeds. Qualitative comparisons of weed detection and stem localization are shown in Fig.~\ref{fig:inference_results} and Fig.~\ref{fig:comp_with_gt_center}.

\subsubsection{Necessity of Object Detection}
With WSD, another learning choice is to regress on weed stem coordinates directly, without the object detection framework. As shown in Tab.~\ref{tab:add_stem_unlabel_det}, the integration of stem regression and weed detection is important for the task of interest. Specifically, adding weed detection reduces the distance error from 4.3306 to 2.4838, demonstrating a significant performance gain. 

\begin{table}[ht]
\centering
\setlength{\tabcolsep}{3mm}
\renewcommand{\arraystretch}{1.5} 
\begin{tabular}{ccc|c}
\hline
Stem Reg. & Det. & Unlabeled & Dist$\downarrow$ \\
\hline
\checkmark & &  & 4.3306 \\
\checkmark & \checkmark &  & 2.4838 \\
\checkmark & \checkmark & \checkmark & \textbf{2.1485} \\
\hline
\end{tabular}
\caption{Analysis on learning components. ``Stem Reg.'' denotes stem regression. ``Det.'' denotes standard object detection. ``Unlabeled'' denotes unlabeled data, which means semi-supervised learning.}
\label{tab:add_stem_unlabel_det}
\end{table}

\subsubsection{Study on Semi-Supervised Learning} 
\begin{figure}[ht]
\centering
\includegraphics[width=\linewidth]{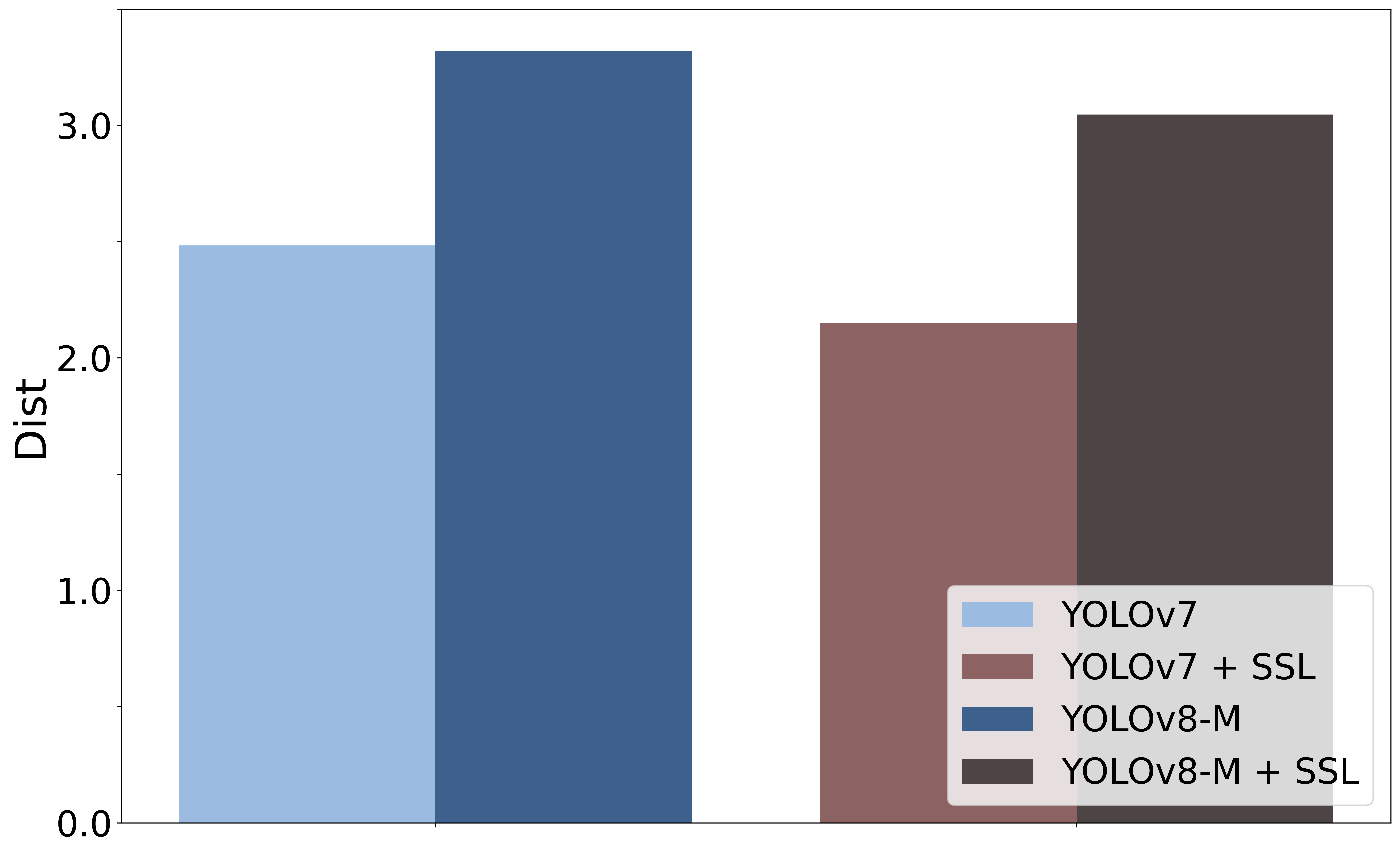}
\caption{Effect of semi-supervised learning. Semi-supervised learning consistently improves the model performance. ``SSL'' denotes semi-supervised learning.} 
\label{compare_weed_bank}
\end{figure}

Using the same setup as above, we evaluated the effectiveness of our semi-supervised learning method, incorporating unlabeled images during training. With semi-supervised learning, the Dist value was further reduced to 2.1485.
Additionally, we replace the detection backbone from YOLOv7 to YOLOv8~\cite{reis2023}. Fig.~\ref{compare_weed_bank} suggests that the extension to semi-supervised learning is model agnostic.

\subsubsection{Ablation Study on Detection Backbone}

Although YOLOv8 generally outperforms YOLOv7 in object detection, it shows lower performance in Fig.~\ref{compare_weed_bank}. We hypothesize that the detection backbone plays a role in this discrepancy. Recent studies~\cite{irjmets2022,baeldung2023,keylabs2023,roboflow2023} indicate that single-stage models, particularly those in the YOLO family, excel in tasks requiring both high-speed inference and competitive accuracy, compared to Faster-RCNN~\cite{ren2016faster} and SSD~\cite{DBLP:conf/eccv/LiuAESRFB16}. We evaluate YOLOv7, YOLOv8, and YOLOv10~\cite{wang2024}, which are anchor-based, anchor-free, and anchor-free, respectively. A comparative analysis is shown in Tab.~\ref{tab:detection_model_compare}.

With stem prediction incorporated, all models achieved a 0\% False Positive rate, ensuring no seedlings were misidentified as weeds. This confirms the effectiveness of our approach in reducing crop damage risk during automated weeding, enhancing both accuracy and safety for young plants. Despite expectations that YOLOv8 and YOLOv10 would perform better, their anchor-free approach resulted in significant drift in stem predictions. In contrast, YOLOv7’s anchor-based method prevents drift by keeping loss calculation confined to the respective grid. The results in Tab.~\ref{tab:detection_model_compare} further validate YOLOv7 as the baseline detection model, balancing computational efficiency and accuracy.

\begin{table}[ht]
\centering
\setlength{\tabcolsep}{1mm}
\renewcommand{\arraystretch}{1.5}
\begin{tabular}{lcccccc}
\hline
Model & \#Param $\downarrow$ & Time(ms)$\downarrow$ & FP$\downarrow$ & Dist$\downarrow$ \\
\hline
YOLOv10-M & \phantom{0}\textbf{26M} & 130.4 & 0 & 3.2691 \\
YOLOv8-M & \phantom{0}\textbf{26M} & 407.3 & 0 & 3.3219\\
YOLOv8-L & \phantom{0}43M & 597.2 & 0 & 3.1473 \\
YOLOv7-X & 142M & 119.5 & 0 & \textbf{2.4426} \\
YOLOv7 & \phantom{0}37M & \phantom{0}\textbf{60.2} & 0 &\textbf{2.4838} \\
\hline
\end{tabular}
\caption{Peformance comparison among different YOLO backbones. ``\#Param'' denotes the number of parameters. ``Time(ms)'' denotes the inference time. ``FP'' denotes the false positive rate.}
\label{tab:detection_model_compare}
\end{table}

\subsubsection{Effect of the Classifier Threshold}
We set a confidence threshold of 0.15 during inference, consistent with most zero-shot object detection models. 
As shown in Tab.~\ref{tab:effect_threshold}, the minimum zero-FP threshold is 0.056.

\begin{table}[ht]
\centering
\begin{tabular}{c|c}
\hline
Threshold & FP \\
\hline
0.150 & 0.000 \\
0.056 & 0.000 \\
0.055 & 0.005 \\
0.050 & 0.020 \\
\hline
\end{tabular}
\caption{Effect of the confidence threshold: raising the threshold can enhance generalization in zero-shot scenarios.}


\label{tab:effect_threshold}
\end{table}

\section{Conclusion}
In this work, we propose an end-to-end pipeline that unifies crop and weed detection and weed stem localization, which shows improved accuracy and reduced energy cost. This study not only serves an empirical study of practical weed recognition, but also poses a promising research direction on intelligent laser weeding.

\section{Acknowledgments}
This work is supported by Shanghai Artificial Intelligence Laboratory.

\bibliography{aaai25}

\begin{thebibliography}{46}
\providecommand{\natexlab}[1]{#1}

\bibitem[{Aamlid et~al.(2021)Aamlid, Hessels{\o}e, Pettersen, and Borchert}]{aamlid2021robo}
Aamlid, T.; Hessels{\o}e, K.~J.; Pettersen, T.; and Borchert, A.~F. 2021.
\newblock ROBO-GOLF: Robotic mowers for better turf quality on golf course fairways and semi-roughs, Results from 2020.
\newblock \emph{NIBIO Rapport}.

\bibitem[{Berthelot et~al.(2019)Berthelot, Carlini, Goodfellow, Papernot, Oliver, and Raffel}]{berthelot2019mixmatch}
Berthelot, D.; Carlini, N.; Goodfellow, I.; Papernot, N.; Oliver, A.; and Raffel, C.~A. 2019.
\newblock Mixmatch: A holistic approach to semi-supervised learning.
\newblock \emph{Advances in neural information processing systems}, 32.

\bibitem[{{Carbon Robotics}(2022)}]{weedlazer}
{Carbon Robotics}. 2022.
\newblock Laserweeder Implement.

\bibitem[{Chen et~al.(2022)Chen, Lu, Li, and Young}]{chen2022performance}
Chen, D.; Lu, Y.; Li, Z.; and Young, S. 2022.
\newblock Performance evaluation of deep transfer learning on multi-class identification of common weed species in cotton production systems.
\newblock \emph{Computers and Electronics in Agriculture}, 198: 107091.

\bibitem[{Dang et~al.(2023)Dang, Chen, Lu, and Li}]{dang2023yoloweeds}
Dang, F.; Chen, D.; Lu, Y.; and Li, Z. 2023.
\newblock YOLOWeeds: A novel benchmark of YOLO object detectors for multi-class weed detection in cotton production systems.
\newblock \emph{Computers and Electronics in Agriculture}, 205: 107655.

\bibitem[{Fawakherji et~al.(2019)Fawakherji, Potena, Bloisi, Imperoli, Pretto, and Nardi}]{fawakherji2019uav}
Fawakherji, M.; Potena, C.; Bloisi, D.~D.; Imperoli, M.; Pretto, A.; and Nardi, D. 2019.
\newblock UAV image based crop and weed distribution estimation on embedded GPU boards.
\newblock In \emph{Computer Analysis of Images and Patterns: CAIP 2019 International Workshops, ViMaBi and DL-UAV, Salerno, Italy, September 6, 2019, Proceedings 18}, 100--108. Springer.

\bibitem[{Hasan et~al.(2021)Hasan, Sohel, Diepeveen, Laga, and Jones}]{hasan2021survey}
Hasan, A.~M.; Sohel, F.; Diepeveen, D.; Laga, H.; and Jones, M.~G. 2021.
\newblock A survey of deep learning techniques for weed detection from images.
\newblock \emph{Computers and Electronics in Agriculture}, 184: 106067.

\bibitem[{Hu et~al.(2024)Hu, Wang, Coleman, Bender, Yao, Zeng, Song, Schumann, and Walsh}]{hu2024deep}
Hu, K.; Wang, Z.; Coleman, G.; Bender, A.; Yao, T.; Zeng, S.; Song, D.; Schumann, A.; and Walsh, M. 2024.
\newblock Deep learning techniques for in-crop weed recognition in large-scale grain production systems: a review.
\newblock \emph{Precision Agriculture}, 25(1): 1--29.

\bibitem[{Jeon, Tian, and Zhu(2011)}]{jeon2011robust}
Jeon, H.~Y.; Tian, L.~F.; and Zhu, H. 2011.
\newblock Robust crop and weed segmentation under uncontrolled outdoor illumination.
\newblock \emph{Sensors}, 11(6): 6270--6283.

\bibitem[{Jiang et~al.(2020)Jiang, Zhang, Qiao, Zhang, Zhang, and Song}]{jiang2020cnn}
Jiang, H.; Zhang, C.; Qiao, Y.; Zhang, Z.; Zhang, W.; and Song, C. 2020.
\newblock CNN feature based graph convolutional network for weed and crop recognition in smart farming.
\newblock \emph{Computers and Electronics in Agriculture}, 174: 105450.

\bibitem[{Joiya(2022)}]{irjmets2022}
Joiya, F. 2022.
\newblock YOLO vs. Faster R-CNN for Object Detection.
\newblock \emph{International Research Journal of Modernization in Engineering Technology and Science (IRJMETS)}.
\newblock Accessed: 2024-08-14.

\bibitem[{Keylabs(2023)}]{keylabs2023}
Keylabs. 2023.
\newblock YOLOv8 vs SSD: Choosing the Right Object Detection Model.
\newblock \emph{Keylabs}.
\newblock Accessed: 2024-08-14.

\bibitem[{Kingma et~al.(2014)Kingma, Mohamed, Jimenez~Rezende, and Welling}]{kingma2014semi}
Kingma, D.~P.; Mohamed, S.; Jimenez~Rezende, D.; and Welling, M. 2014.
\newblock Semi-supervised learning with deep generative models.
\newblock \emph{Advances in neural information processing systems}, 27.

\bibitem[{Liu et~al.(2016)Liu, Anguelov, Erhan, Szegedy, Reed, Fu, and Berg}]{DBLP:conf/eccv/LiuAESRFB16}
Liu, W.; Anguelov, D.; Erhan, D.; Szegedy, C.; Reed, S.~E.; Fu, C.; and Berg, A.~C. 2016.
\newblock {SSD:} Single Shot MultiBox Detector.
\newblock In \emph{European Conference on Computer Vision}, 21--37.

\bibitem[{Long, Shelhamer, and Darrell(2015)}]{long2015fully}
Long, J.; Shelhamer, E.; and Darrell, T. 2015.
\newblock Fully convolutional networks for semantic segmentation.
\newblock In \emph{Proceedings of the IEEE conference on computer vision and pattern recognition}, 3431--3440.

\bibitem[{Lu(2023)}]{lu2023cottonweeddet12}
Lu, Y. 2023.
\newblock CottonWeedDet12: A 12-class weed dataset of cotton production systems for benchmarking AI models for weed detection.
\newblock \emph{Zenodo}.

\bibitem[{Nanos(2023)}]{baeldung2023}
Nanos, G. 2023.
\newblock Object Detection: SSD Vs. YOLO.
\newblock \emph{Baeldung on Computer Science}.
\newblock Accessed: 2024-08-14.

\bibitem[{Nasiri et~al.(2022)Nasiri, Omid, Taheri-Garavand, and Jafari}]{nasiri2022deep}
Nasiri, A.; Omid, M.; Taheri-Garavand, A.; and Jafari, A. 2022.
\newblock Deep learning-based precision agriculture through weed recognition in sugar beet fields.
\newblock \emph{Sustainable computing: Informatics and systems}, 35: 100759.

\bibitem[{Olsen et~al.(2019)Olsen, Konovalov, Philippa, Ridd, Wood, Johns, Banks, Girgenti, Kenny, Whinney et~al.}]{olsen2019deepweeds}
Olsen, A.; Konovalov, D.~A.; Philippa, B.; Ridd, P.; Wood, J.~C.; Johns, J.; Banks, W.; Girgenti, B.; Kenny, O.; Whinney, J.; et~al. 2019.
\newblock DeepWeeds: A multiclass weed species image dataset for deep learning.
\newblock \emph{Scientific Reports}, 9(1): 2058.

\bibitem[{Parra et~al.(2020)Parra, Marin, Yousfi, Rinc{\'o}n, Mauri, and Lloret}]{parra2020edge}
Parra, L.; Marin, J.; Yousfi, S.; Rinc{\'o}n, G.; Mauri, P.~V.; and Lloret, J. 2020.
\newblock Edge detection for weed recognition in lawns.
\newblock \emph{Computers and Electronics in Agriculture}, 176: 105684.

\bibitem[{Pirchio et~al.(2018)Pirchio, Fontanelli, Frasconi, Martelloni, Raffaelli, Peruzzi, Caturegli, Gaetani, Magni, Volterrani et~al.}]{pirchio2018}
Pirchio, M.; Fontanelli, M.; Frasconi, C.; Martelloni, L.; Raffaelli, M.; Peruzzi, A.; Caturegli, L.; Gaetani, M.; Magni, S.; Volterrani, M.; et~al. 2018.
\newblock Autonomous rotary mower versus ordinary reel mower—effects of cutting height and nitrogen rate on Manila grass turf quality.
\newblock \emph{HortTechnology}, 28(4): 509--515.

\bibitem[{Rahman, Lu, and Wang(2023)}]{rahman2023performance}
Rahman, A.; Lu, Y.; and Wang, H. 2023.
\newblock Performance evaluation of deep learning object detectors for weed detection for cotton.
\newblock \emph{Smart Agricultural Technology}, 3: 100126.

\bibitem[{Reis et~al.(2023)Reis, Kupec, Hong, and Daoudi}]{reis2023}
Reis, D.; Kupec, J.; Hong, J.; and Daoudi, A. 2023.
\newblock Real-time flying object detection with YOLOv8.
\newblock \emph{arXiv preprint arXiv:2305.09972}.

\bibitem[{Ren et~al.(2016)Ren, He, Girshick, and Sun}]{ren2016faster}
Ren, S.; He, K.; Girshick, R.; and Sun, J. 2016.
\newblock Faster R-CNN: Towards real-time object detection with region proposal networks.
\newblock \emph{IEEE Transactions on Pattern Analysis and Machine Intelligence}, 39(6): 1137--1149.

\bibitem[{Roboflow(2023)}]{roboflow2023}
Roboflow. 2023.
\newblock YOLOv8 vs. Faster R-CNN: Compared and Contrasted.
\newblock Accessed: 2024-08-14.

\bibitem[{Sharpe et~al.(2020)Sharpe, Schumann, Yu, and Boyd}]{sharpe2020vegetation}
Sharpe, S.~M.; Schumann, A.~W.; Yu, J.; and Boyd, N.~S. 2020.
\newblock Vegetation detection and discrimination within vegetable plasticulture row-middles using a convolutional neural network.
\newblock \emph{Precision Agriculture}, 21: 264--277.

\bibitem[{Sportelli et~al.(2020)Sportelli, Pirchio, Fontanelli, Volterrani, Frasconi, Martelloni, Caturegli, Gaetani, Grossi, Magni et~al.}]{sportelli2020autonomous}
Sportelli, M.; Pirchio, M.; Fontanelli, M.; Volterrani, M.; Frasconi, C.; Martelloni, L.; Caturegli, L.; Gaetani, M.; Grossi, N.; Magni, S.; et~al. 2020.
\newblock Autonomous mowers working in narrow spaces: A possible future application in agriculture?
\newblock \emph{Agronomy}, 10(4): 553.

\bibitem[{Steininger et~al.(2023)Steininger, Trondl, Croonen, Simon, and Widhalm}]{steininger2023cropandweed}
Steininger, D.; Trondl, A.; Croonen, G.; Simon, J.; and Widhalm, V. 2023.
\newblock The cropandweed dataset: A multi-modal learning approach for efficient crop and weed manipulation.
\newblock In \emph{Proceedings of the IEEE/CVF Winter Conference on Applications of Computer Vision}, 3729--3738.

\bibitem[{Sudars et~al.(2020)Sudars, Jasko, Namatevs, Ozola, and Badaukis}]{sudars2020dataset}
Sudars, K.; Jasko, J.; Namatevs, I.; Ozola, L.; and Badaukis, N. 2020.
\newblock Dataset of annotated food crops and weed images for robotic computer vision control.
\newblock \emph{Data in brief}, 31: 105833.

\bibitem[{Tanveer et~al.(2003)Tanveer, Chaudhry, Ayub, and Ahmad}]{tanveer2003}
Tanveer, A.; Chaudhry, N.; Ayub, M.; and Ahmad, R. 2003.
\newblock Effect of cultural and chemical weed control methods on weed population and yield of cotton.
\newblock \emph{Pak. J. Bot}, 35(2): 161--166.

\bibitem[{Tao and Wei(2022)}]{tao2022hybrid}
Tao, T.; and Wei, X. 2022.
\newblock A hybrid CNN--SVM classifier for weed recognition in winter rape field.
\newblock \emph{Plant Methods}, 18(1): 29.

\bibitem[{{United Nations}(2023)}]{UnitedNations2023b}
{United Nations}. 2023.
\newblock Leave no one behind.
\newblock \url{https://unsdg.un.org/2030-agenda/universal-values/leave-no-one-behind}.

\bibitem[{Veeranampalayam~Sivakumar et~al.(2020)Veeranampalayam~Sivakumar, Li, Scott, Psota, J.~Jhala, Luck, and Shi}]{veeranampalayam2020comparison}
Veeranampalayam~Sivakumar, A.~N.; Li, J.; Scott, S.; Psota, E.; J.~Jhala, A.; Luck, J.~D.; and Shi, Y. 2020.
\newblock Comparison of object detection and patch-based classification deep learning models on mid-to late-season weed detection in UAV imagery.
\newblock \emph{Remote Sensing}, 12(13): 2136.

\bibitem[{Wang et~al.(2024)Wang, Chen, Liu, Chen, Lin, Han, and Ding}]{wang2024}
Wang, A.; Chen, H.; Liu, L.; Chen, K.; Lin, Z.; Han, J.; and Ding, G. 2024.
\newblock Yolov10: Real-time end-to-end object detection.
\newblock \emph{arXiv preprint arXiv:2405.14458}.

\bibitem[{Wang, Bochkovskiy, and Liao(2023)}]{wang2023}
Wang, C.-Y.; Bochkovskiy, A.; and Liao, H.-Y.~M. 2023.
\newblock YOLOv7: Trainable bag-of-freebies sets new state-of-the-art for real-time object detectors.
\newblock In \emph{Proceedings of the IEEE/CVF conference on computer vision and pattern recognition}, 7464--7475.

\bibitem[{{WeLASER}(2023)}]{LaserW2023}
{WeLASER}. 2023.
\newblock WeLASER Implement.

\bibitem[{Wu et~al.(2021)Wu, Chen, Zhao, Kang, and Ding}]{wu2021review}
Wu, Z.; Chen, Y.; Zhao, B.; Kang, X.; and Ding, Y. 2021.
\newblock Review of weed detection methods based on computer vision.
\newblock \emph{Sensors}, 21(11): 3647.

\bibitem[{Xu et~al.(2021)Xu, Zhang, Hu, Wang, Wang, Wei, Bai, and Liu}]{xu2021end}
Xu, M.; Zhang, Z.; Hu, H.; Wang, J.; Wang, L.; Wei, F.; Bai, X.; and Liu, Z. 2021.
\newblock End-to-end semi-supervised object detection with soft teacher.
\newblock In \emph{Proceedings of the IEEE/CVF international conference on computer vision}, 3060--3069.

\bibitem[{You, Liu, and Lee(2020)}]{you2020dnn}
You, J.; Liu, W.; and Lee, J. 2020.
\newblock A DNN-based semantic segmentation for detecting weed and crop.
\newblock \emph{Computers and Electronics in Agriculture}, 178: 105750.

\bibitem[{Yu et~al.(2019)Yu, Schumann, Cao, Sharpe, and Boyd}]{yu2019weed}
Yu, J.; Schumann, A.~W.; Cao, Z.; Sharpe, S.~M.; and Boyd, N.~S. 2019.
\newblock Weed detection in perennial ryegrass with deep learning convolutional neural network.
\newblock \emph{Frontiers in Plant Science}, 10: 1422.

\bibitem[{Zhai et~al.(2019)Zhai, Oliver, Kolesnikov, and Beyer}]{zhai2019s4l}
Zhai, X.; Oliver, A.; Kolesnikov, A.; and Beyer, L. 2019.
\newblock S4l: Self-supervised semi-supervised learning.
\newblock In \emph{Proceedings of the IEEE/CVF international conference on computer vision}, 1476--1485.

\bibitem[{Zhang et~al.(2024)Zhang, Cao, Zhou, and Currie}]{zhang2024laser}
Zhang, H.; Cao, D.; Zhou, W.; and Currie, K. 2024.
\newblock Laser and optical radiation weed control: a critical review.
\newblock \emph{Precision Agriculture}, 1--25.

\bibitem[{Zhang, Zhong, and Zhou(2023)}]{zhang2023}
Zhang, H.; Zhong, J.-X.; and Zhou, W. 2023.
\newblock Precision optical weed removal evaluation with laser.
\newblock In \emph{CLEO: Applications and Technology}, JW2A--145. Optica Publishing Group.

\bibitem[{Zhang et~al.(2022)Zhang, Zhang, Wu, and Sun}]{zhang2022segmentation}
Zhang, L.; Zhang, Z.; Wu, C.; and Sun, L. 2022.
\newblock Segmentation algorithm for overlap recognition of seedling lettuce and weeds based on SVM and image blocking.
\newblock \emph{Computers and Electronics in Agriculture}, 201: 107284.

\bibitem[{Zhang(1996{\natexlab{a}})}]{zhang1996}
Zhang, Z. 1996{\natexlab{a}}.
\newblock Developing chemical weed control and attaching importance to integrated weed management.
\newblock In \emph{Proceedings of the National Symposium of IPM in China, 1996}. China Agricultural Science and Technology Press.

\bibitem[{Zhang(1996{\natexlab{b}})}]{Zhang1996b}
Zhang, Z. 1996{\natexlab{b}}.
\newblock \emph{Weeds and their control in cotton fields}, volume~2, 1345--1349.
\newblock Beijing: China Agriculture Press.

\end{thebibliography}

\end{document}